# Lifelong learning challenges in the era of artificial intelligence: a computational thinking perspective.

Margarida Romero
Université Côte d'Azur, France

https://orcid.org/0000-0003-3356-8121

**Abstract**
The rapid advancement of artificial intelligence (AI) has brought significant challenges to the education and workforce skills required to take advantage of AI for human-AI collaboration in the workplace. As AI continues to reshape industries and job markets, the need to define how AI literacy can be considered in lifelong learning has become increasingly critical (Cetindamar et al., 2022; Laupichler et al., 2022; Romero et al., 2023). Like any new technology, AI is the subject of both hopes and fears, and what it entails today presents major challenges (Cugurullo & Acheampong, 2023; Villani et al., 2018). It also raises profound questions about our own humanity. Will the machine surpass the intelligence of the humans who designed it? What will be the relationship between so-called AI and our human intelligences? How could human-AI collaboration be regulated in a way that serves the Sustainable Development Goals (SDGs)? This paper provides a review of the challenges of lifelong learning in the era of AI from a computational thinking, critical thinking, and creative competencies perspective, highlighting the implications for management and leadership in organizations.

*Keywords:* artificial intelligence, computational thinking, creativity, critical thinking, lifelong learning.

**Introduction**
Critical thinking is the most transversal competency for the education of the 21st century (Henriksen et al., 2016; Romero, Lille, et al., 2017). Critical thinking is a human-specific competency that requires sensible criteria based on a certain cultural context and interpersonal human relationships (Romero, 2018; Sternberg & Halpern, 2020). Current AI tools require human effort to correct the bias generated by the corpus of data in a way that permits the generation of information that is sensitive to human qualities such as human rights and equality. In this paper, we provide a critical perspective of artificial intelligence (AI) and discuss the importance of developing computational thinking (Lafuente Martínez et al., 2022; Romero, Lepage, et al., 2017; Wing, 2006), creativity (Alexandre et al., 2022; Henriksen et al., 2016; Isaac et al., 2022) and critical thinking (Lucas & Spencer, 2017; Sternberg & Halpern, 2020) for supporting lifelong learning competencies for a better understanding and developing the human-AI activities which can support the workforce in their professional but also personal activities.

**Human intelligence and artificial intelligence**
We start by questioning the term "AI." Is the term "intelligence" relevant to computer applications based on machine learning in particular? The aim of these algorithms is to develop systems capable of capturing, processing, and reacting to (massive) information according to mechanisms that adapt to the context or data to maximize the chances of achieving the objectives defined for the system (Alexandre et al., 2021). This behavior, which may seem "intelligent", is created by human computer scientists and has limitations related to the current human capacity to define efficient machine learning systems and to the availability of massive data for the systems to adapt. It has been found that these systems perform better than humans on specific tasks such as sound and image recognition or tests such as the Stanford Question Answering Dataset (SQuAD) dataset (Tang & Kejriwal, 2023). Does having a better reading test mean being able to understand, in a human and intelligent sense, the text read? The statistical ability to identify answers may seem intelligent, but there is no evidence that it is in the human critical and creative sense.

When, in the 1950s, Turing proposed a test based on a purely linguistic confrontation between a human and another agent, which could be a machine or another human, he was not targeting the intelligence of the machine, but the intelligence we might attribute to it. If the human judges that he is interacting with a human agent and not a machine, the artificial intelligence test is considered successful. And can we be satisfied with a good ability to respond to a human conversation to consider a machine intelligent? (Alexandre et al., 2020).

**Defining human intelligence**

If we think of intelligence as the ability to learn (Beckmann, 2006) and learning as adaptation to context (Piaget, 1953), it would be possible to consider systems that are able to improve their adaptation to context from data collection and processing as intelligent. However, if we consider intelligence as "the ability to solve problems or create solutions that are valuable in a given socio-cultural context" (Gardner & Hatch, 1989, p. 5), under a diverse and dynamic approach, it is more difficult to consider that a system, however adaptive and massively fed with data, can make metacognitive judgements about its process and products in relation to a given socio-cultural context. Gardner and Hatch's definition of human intelligence is very similar to that of creativity as the process of designing a solution that is deemed new, innovative and relevant in relation to the specific context of the problem situation (Romero, Lille, et al., 2017). Intelligence is therefore not the ability to perform according to pre-established or predictable rules (including with adaptation mechanisms or machine learning on data), but rather the ability to create something new by demonstrating a faculty of sensitivity and adaptation to the socio-cultural context and empathy on an intra- and inter-psychological level with the different actors. This implies an understanding of human and socio-historical nature in order to be able to judge one's own process and creation autonomously.

If we adopt this second critical and creative approach to intelligence, we should be cautious about using the term AI for solutions that 'only' adapt according to predefined mechanisms that cannot generate self-reflexive value judgments nor socio-cultural perspectives. Machine learning systems that are labelled AI may be very good at basing themselves on very elaborate models fed with massive data, but they are not "intelligent" in the critical and creative way that humans are. For example, my phone can learn to recognize the words I dictate vocally, even if I have an accent, which it will infer the more I use the system. However, to attribute real intelligence to a voice dictation is a subjective projection, i.e., just a *belief* (Alexandre et al., 2023).

**Developing critical thinking, creativity, and computational thinking**

We can also question the "intelligence of AI" in relation to the critical thinking that characterizes human intelligence. In the context of the #CoCreaTIC project, we define critical thinking as the ability to develop independent critical thinking, which allows for the analysis of ideas, knowledge, and processes in relation to a system of values and own judgements (Romero, Lille, et al., 2017). It is responsible thinking that is based on criteria and sensitive to context and others. On the other hand, if we think of algorithmic learning systems, and the politically incorrect results they have produced in the face of images and textual responses that can be labelled as discriminatory, we should neither fear, nor condemn, nor accept, this result, because it has no moral value. The most likely explanation is that by 'learning' data from humans, the mechanism highlights racist and sexist elements; there is no value system of its own. Here, this so-called AI does not have responsible thinking but exacerbates certain deviations that humans are capable of producing but also of limiting and correcting by their criteria and sensitivity to others.

Supporting critical thinking and creativity are essential for lifelong learning in the AI era as a way of preparing citizens for the challenges of what is made possible by these algorithms. Digital literacy, especially in critical, creative, and participatory approaches, can also help to develop a relationship with computing that allows citizens to demystify AI, develop an ethical requirement, and adopt an informed attitude in order to accept or not what will be used in their personal, social, or professional activities. For acculturating lifelong learners to the fundamentals of AI, developing computational thinking (Lafuente Martínez et al., 2022) is also a crucial aspect of lifelong learning. Computational thinking involves problem-solving, critical thinking, and decision-making skills that are essential for navigating the complex and dynamic world of AI. For these reasons, the development of computational thinking competency is also an asset that complements the need to develop critical and creative thinking skills for understanding AI and developing human-AI collaboration in education and the workplace.

**The lever of computational thinking for supporting human-AI collaboration**

In 2006, Wing described "computational thinking" as the ability to use computer processes to solve problems in any domain. To develop it, learners (from kindergarten onwards, and at all ages) can combine the learning of computational concepts and processes that are the subject of 'digital literacy' (object, attribute, method, design pattern, etc.) with a creative problem-solving approach that uses computational concepts and processes (Romero, Lepage, et al., 2017). Projects such as Class'Code in France or CoCreaTic in Canada have developed resources and a community to support this approach, in which it is not about learning 'coding' (in the sense of coding with a computer language) step by step, but about solving problems creatively and in a way that is sensitive to the context of the problem. In other words, going beyond coding allows us to anchor ourselves in a broader approach to creative programming (Menon et al., 2019; Romero et al., 2016). This engages learners because it is a critical and creative problem-solving process that draws on computer science concepts and processes. It is not about coding for coding's sake, or writing lines of code one after the other, but about developing an approach to solving complex problems that engages in a reflective and empathetic analysis of the situation, its representation, and the operationalization of a solution that takes advantage of the metacognitive strategies associated with computational thinking.

Computational thinking is presented by Wing as a set of attitudes and knowledge that are universally applicable, beyond the use of machines; "this set of intellectual tools includes, for example, the ability to name objects in a relevant way and to make their type or category explicit in order to manipulate them correctly, to master the complexity of a large problem or system in a mechanical way (Wing, 2009, para. 3) . The term 'computational thinking' is used to show that the aim is not only to introduce students to programming but also to enable them to take a step back from the digital world and to position the learning of computer science as a cross-curricular skill that draws on the digital world to develop thinking strategies. Computational thinking is thus intended to be emancipatory, with the aim of helping to form enlightened citizens who can be both critically minded actors and also have the capacity to understand and (co)create with the "panoply of digital tools" that Wing evokes when talking about computational thinking.

The development of a critical and creative approach to digital technology through computational thinking allows learners to move beyond a user posture that might perceive AI as a black box full of mysteries, dangers, or unlimited hopes. The understanding of the challenges of analyzing problems in relation to specific socio-cultural contexts (e.g., migration issues) is a way of seeing computer science as both a science and a technology that will allow us, starting from the limits and constraints of our models of a problem, to try to give answers that will be fed by increasingly massive data, without being able to be considered relevant or valuable without the involvement of human judgement.

Computational thinking competency development can serve for addressing the challenges of human-AI collaboration and for understanding and navigating the complex world of AI, as they enable individuals to analyze problems, develop strategies, and make informed decisions in the context of rapidly changing technological landscapes. Furthermore, computational thinking can promote critical thinking and creativity, which are vital for effectively managing and leading organizations that are influenced by AI technologies. The implications of computational thinking for management and leadership in organizations are essential at different levels of the workforce. Leaders need to foster a culture of lifelong learning that encourages all the workforce to develop computational thinking competencies to better consider the way to incorporate AI-human improvements in their work. This may involve providing training and development opportunities, creating supportive learning environments, and promoting collaboration to address human-AI challenges. For this objective, Romero et al. (2022) propose a Techno-creative Problem-Solving (TCPS) framework for the development of computational thinking through creative programming in lifelong learning.

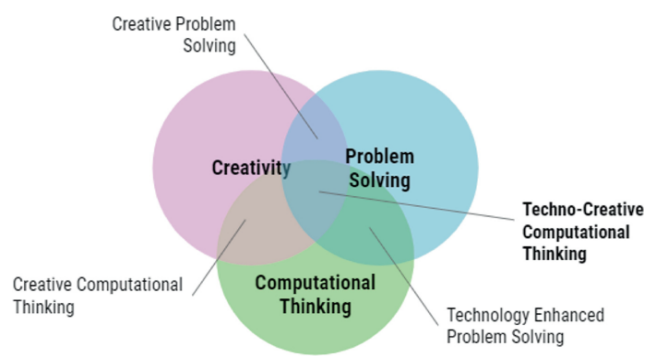

Figure 1. Techno-creative Problem-Solving (TCPS) framework

**Human-AI collaboration in HRM**

The literature review draws on several studies that have explored the challenges and opportunities of AI in different contexts. Budhwar et al. (2022) conducted a review and research agenda on the challenges and opportunities of AI for international human resource management (HRM). They highlighted the need for HRM to adapt to the changes brought by AI, including the development of new skills and competencies for managing AI technologies in an international context. Fügener et al. (2022) investigated the cognitive challenges in human-artificial intelligence collaboration and explored the path towards productive delegation. They emphasized the importance of understanding the cognitive capabilities and limitations of both humans and AI systems in order to optimize their collaboration. Arslan et al. (2021) discussed the challenges and potential HRM strategies for managing the interaction between artificial intelligence and human workers at the team level. They highlighted the need for HRM to address the social, ethical, and psychological challenges associated with AI integration in teams.

**Discussion**

This paper highlights the importance of computational thinking, critical thinking, and creativity in the context of lifelong learning and their implications for management and leadership in organizations. Our society needs to face the emergence of AI with a more critical and creative education. This lifelong learning challenge presents significant challenges for the workforce; however, by embracing computational thinking, critical thinking, and creativity as key competencies, individuals and organizations can effectively address these challenges and thrive in the rapidly evolving technological landscape. For informed citizenship in the digital age, we need to continue to sharpen our critical, creative, and collaborative problem-solving skills while adding a new string to our bow: the development of computational thinking.